\documentclass[letterpaper,10pt,conference]{ieeeconf}

\IEEEoverridecommandlockouts
\usepackage{graphicx}
\usepackage{cite}
\usepackage{amsmath,amssymb}
\usepackage{booktabs}
\usepackage{multirow}
\usepackage{tabularx}
\usepackage{array}
\usepackage{microtype}
\usepackage{float}
\usepackage{newtxtext,newtxmath}

\title{MIM-VLA: Learning Physical Interaction Representations from Gripper
Motor Feedback}

\author{\begin{tabular}{c}
Jaeyoung Lee\textsuperscript{*}, Jiyeon Koo\textsuperscript{*}, Taehwa Kim, Yerin Cha, and Andrew Jaeyong Choi\textsuperscript{\textdagger}\\[-1pt]
\small School of Computing, iRASC Lab., Gachon University, Seongnam, Republic of Korea\\[-1pt]
\footnotesize \texttt{\{ljy0111, halo1225, tempteru, chayerin123, andrewjchoi\}@gachon.ac.kr}\\[-1pt]
\footnotesize \textsuperscript{*}Equal contribution as co-first authors. \textsuperscript{\textdagger}Corresponding author.
\end{tabular}}

\begin{document}
\maketitle
\thispagestyle{empty}
\pagestyle{empty}


\begin{abstract}
Vision-language-action (VLA) policies infer grasp actions primarily from
visual observations and robot state, but do not explicitly represent the
physical response observed after contact. We present \textbf{MIM-VLA}, a
motor-feedback-based architecture that encodes recent gripper current,
position, velocity, and signal validity as a 128-dimensional interaction
token $\boldsymbol{\tau}$. A motor-only Motor Interaction Module (MIM) is pretrained with
human-reviewed contact and interaction-phase labels and then conditions only
the gripper-action pathway of SmolVLA; arm actions and the position-control
interface remain unchanged. The same token supports the MEM selector VLM that
compares candidate interactions and produces evidence-conditioned selections
and explanations. We evaluate MIM-VLA in three real-world settings: comparing
the interaction resistance of visually different objects, disambiguating
visually similar real and replica objects through active probing, and gently
grasping fragile objects, including held-out instances. Across 13 object pairs,
MIM-VLA selects the higher-resistance object in 75.0\% of trials, compared with
48.8\% for the SmolVLA baseline. For the evaluated tasks, the approach uses
motor feedback already available from the gripper and does not require an
additional tactile array, force--torque sensor, calibrated force estimate, or
direct current control.
\end{abstract}

\section{Introduction}
\label{sec:introduction}

Vision-language-action (VLA) models connect images and natural-language
instructions to robot actions \cite{openvla,smolvla}. Their visual-language
priors support object recognition and task planning, yet appearance does not
fully specify how an object will respond after contact. Visually similar
objects may have different compliance, while fragile objects of different
sizes cannot be safely grasped by replaying one closing position. Joint
position alone also cannot distinguish free closing, contact, stable holding,
and continued compression.

Force- and tactile-aware policies address this limitation using dedicated
sensors or force-aware control \cite{forcevla,tactilevla,biact}. We instead
ask whether actuator telemetry already exposed by a smart gripper can provide
useful physical grounding. Motor current is not calibrated force and depends
on motion and actuator dynamics; consequently, we model its temporal relation
with measured position, velocity, and signal validity rather than applying an
instantaneous current threshold.

\begin{figure}[t!]
    \centering
    \includegraphics[width=0.88\columnwidth]{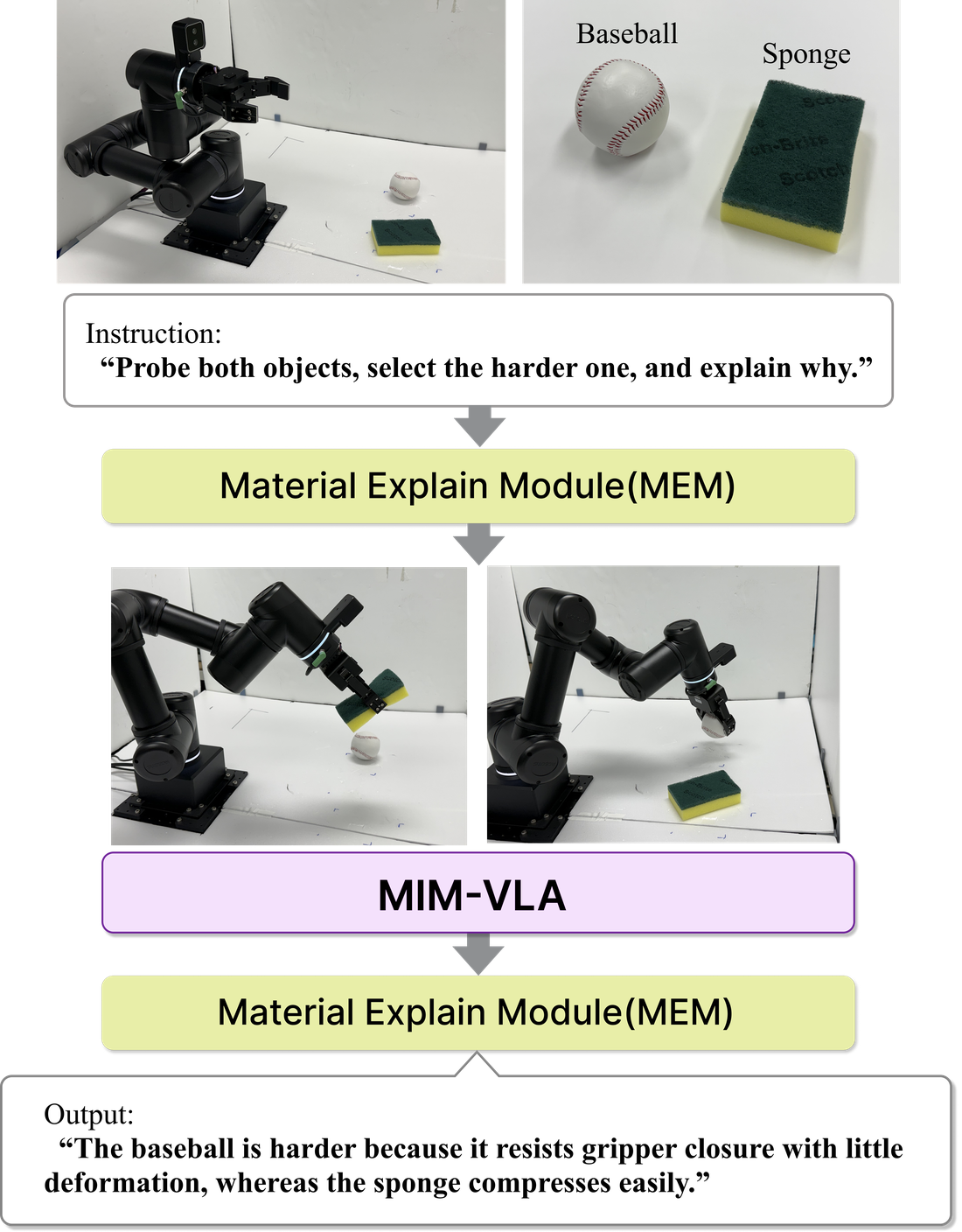}
    \caption{Interaction-grounded candidate selection. The robot probes both
    objects, compares their motor responses, selects the object satisfying the
    requested physical property, and explains its decision.}
    \label{fig:material_selection}
\end{figure}

\begin{figure*}[t]
    \centering
    \includegraphics[width=\textwidth]{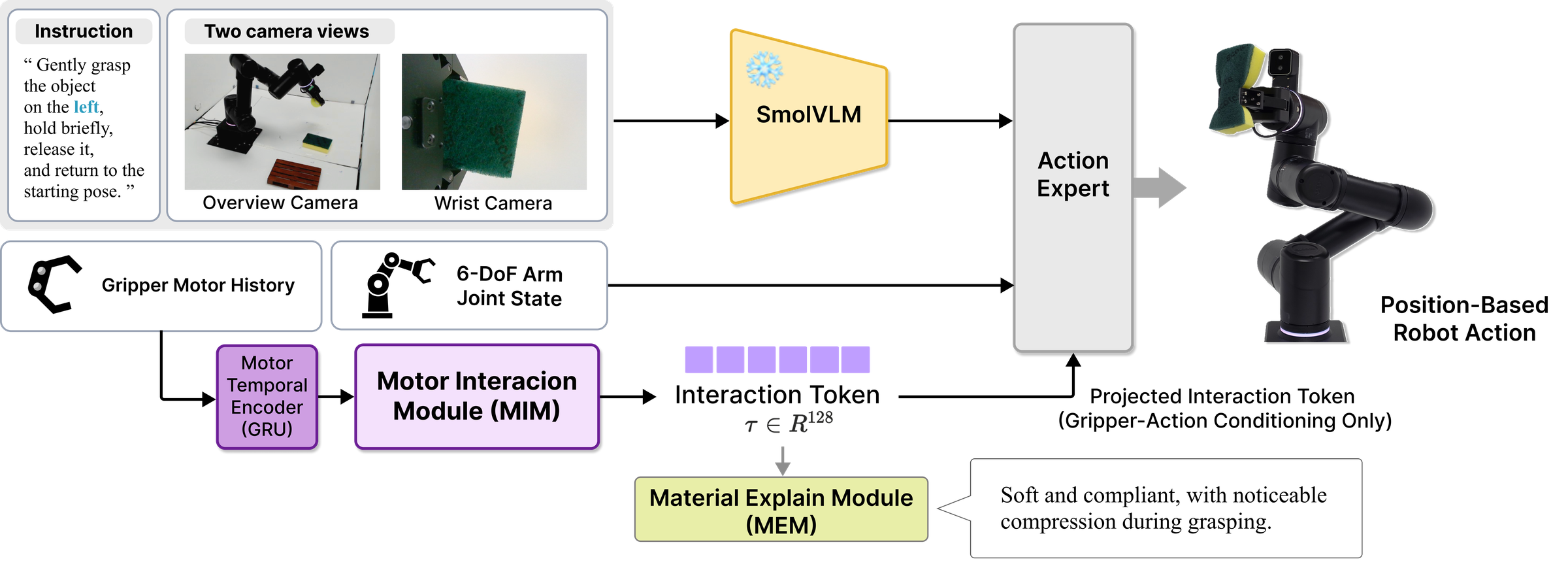}
    \caption{MIM-VLA architecture. Multi-view observations and language are
    processed by the frozen SmolVLM backbone. The MIM interaction token
    conditions gripper-action prediction and supports evidence-conditioned
    comparison through the Material Explain Module (MEM).}
    \label{fig:mim_vla_architecture}
\end{figure*}

\begin{figure}[t]
    \centering
    \includegraphics[width=\columnwidth]{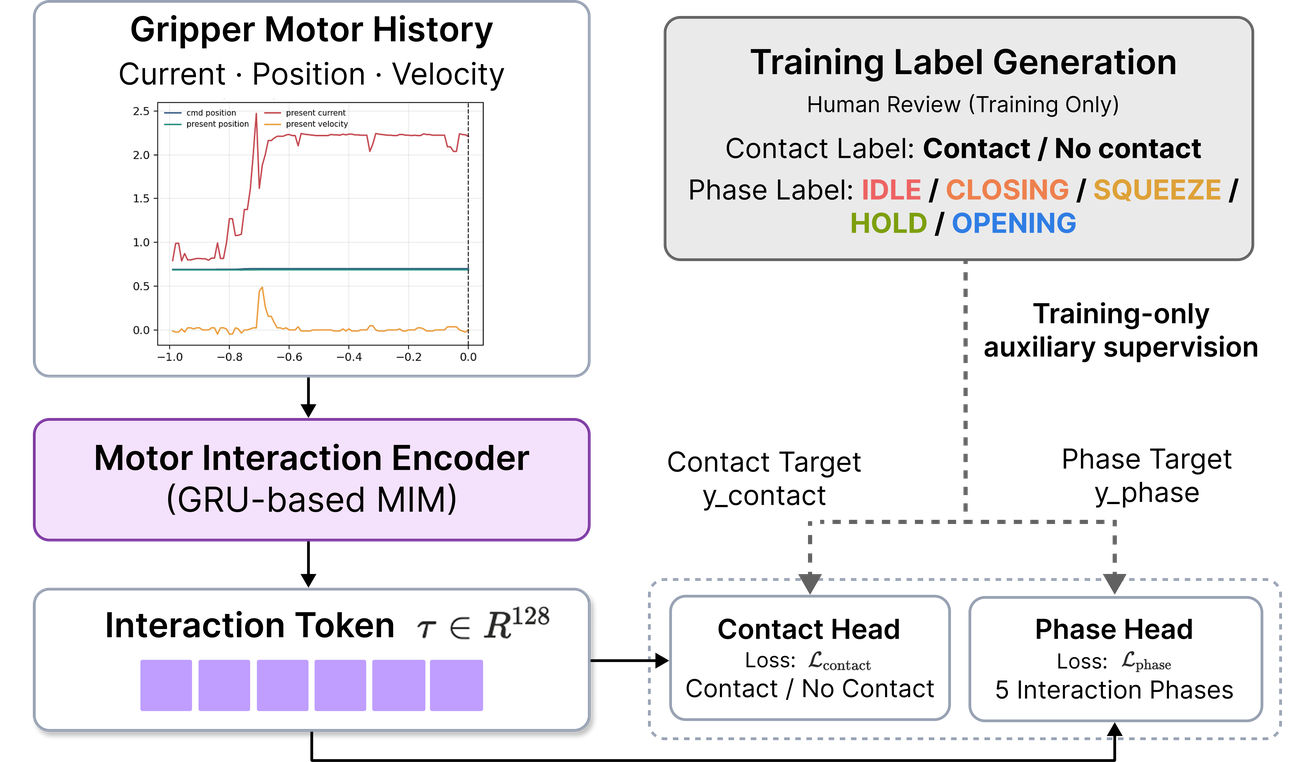}
    \caption{Motor-only MIM pretraining. Human-reviewed contact and phase
    targets supervise auxiliary heads during training; video and the commanded
    position shown for annotation context are visualization only and are
    excluded from the MIM encoder input.}
    \label{fig:mim_pretraining}
\end{figure}

MIM-VLA, shown in Fig.~\ref{fig:mim_vla_architecture}, uses a GRU-based Motor
Interaction Module (MIM) to encode recent gripper history as an interaction
token $\boldsymbol{\tau}\in\mathbb{R}^{128}$. MIM is pretrained with
binary contact and five interaction-phase labels---\textsc{idle},
\textsc{closing}, \textsc{squeeze}, \textsc{hold}, and \textsc{opening}.
Video supports label generation but is excluded from the motor-only encoder
(Fig.~\ref{fig:mim_pretraining}). During policy learning, a lightweight
projector injects the token only into a dedicated gripper pathway. The
SmolVLM backbone and arm pathway are preserved, and the final outputs remain
position commands rather than current or force commands.

For these interaction states, position and velocity describe aperture and
closure progression, while current supplies the corresponding motor-load
response. MIM remains motor-only to avoid appearance shortcuts; vision stays
in the SmolVLM pathway and is not claimed to be replaced by telemetry.

The same representation supports interaction-level reasoning through the
Material Explain Module (MEM). Candidate-specific token histories are
summarized for a selector VLM, which compares measured interaction responses,
selects a candidate, and renders a short explanation; low-level execution
remains with the action policy. MEM demonstrates reuse of the same physical
token beyond low-level control. We evaluate three capabilities: (i) comparing
relative interaction resistance across visually different objects, (ii)
probing visually similar real and replica objects to identify the more
resistant candidate, as illustrated in
Fig.~\ref{fig:material_selection}, and (iii) minimally compressive grasping of
seen and held-out fragile objects. Comparisons include vanilla SmolVLA,
raw-motor conditioning, and a TA-VLA-inspired temporal-current baseline under
the same position-action interface.

Our contributions are:
\begin{itemize}
    \item a motor-only interaction representation pretrained from temporal
    gripper telemetry and contact/phase supervision, without calibrated force
    sensing;
    \item a gripper-specific integration that preserves the pretrained
    visual-language backbone, arm pathway, and position-control interface; and
    \item a shared token interface for closed-loop gripper adaptation and
    higher-level evidence-conditioned comparison and explanation through MEM,
    evaluated in three complementary real-world settings.
\end{itemize}

\section{Related Work}
\label{sec:related_work}

\subsection{Vision-Language-Action Models}

Vision-language-action models translate pretrained visual-language
representations into robot actions. OpenVLA presents an open-source generalist
policy pretrained on diverse robot demonstrations and supports efficient
adaptation to new settings \cite{openvla}.~More recent VLAs employ continuous
generative action models: $\pi_0$ adopts flow matching for general robot
control \cite{pi0}, while GR00T N1 couples a vision-language module with a
diffusion-transformer action module for generalist humanoid manipulation
\cite{grootn1}. FLARE augments generative policy learning with future latent
representation alignment, encouraging the policy to model consequences in a
control-relevant feature space \cite{flare}. SmolVLA combines a compact
flow-matching architecture with asynchronous inference for efficient
deployment on affordable hardware \cite{smolvla}. These methods learn rich
representations for action prediction, but they do not explicitly preserve
object-associated evidence acquired through prior physical probing. Building
on SmolVLA, our approach summarizes gripper motor-response histories as
interaction tokens that can be associated with explored objects and used to
condition subsequent gripper control.

\subsection{Physical Interaction for Robot Manipulation}

Recent work incorporates physical feedback at different levels of the
perception--action hierarchy. Among the approaches most closely related to
ours, ForceVLA fuses real-time force--torque measurements with
vision-language features through a force-aware mixture-of-experts during
action decoding \cite{forcevla}. TA-VLA studies torque-aware adapters,
temporal torque history, and auxiliary torque prediction for integrating
physical feedback into existing VLA architectures \cite{tavla}. ForceVLA2
introduces force prompts for task-level reasoning and combines them with
hybrid force--position control \cite{forcevla2}. Tactile approaches align
touch with vision and language for contact-rich insertion, as in VTLA
\cite{vtla}, or use tactile feedback at both the planning and action-refinement
levels, as in VLA-Touch \cite{vlatouch}. Bi-ACT instead learns position,
velocity, and force behavior from demonstrations collected with bilateral
control \cite{biact}.

These approaches primarily use explicit force or tactile signals to improve
ongoing planning and control, or require a force-aware demonstration and
execution interface. In contrast, MIM operates on the temporal current,
position, and velocity signals already exposed by the gripper. Wrist video is
used only to generate and review auxiliary pretraining labels; it is not an
input to the motor-only MIM. We do not interpret current as calibrated force.
Instead, MIM encodes the joint evolution of motor load and finger motion into
an interaction token, allowing the original position-action interface to be
retained.

\subsection{Interaction Grounding with Vision-Language Models}

Vision-language models connect visual observations with linguistic
descriptions. Qwen2.5-VL supports spatial object localization and temporal
event localization in long videos \cite{qwen25vl}, while SmolVLM provides a
compact architecture for image and video understanding \cite{smolvlm}.
Visual evidence alone, however, may not reveal the physical response produced
during grasping. PaLM-E demonstrates that continuous robot-state estimates can
be interleaved with vision and language to support embodied reasoning
\cite{palme}. Other work uses VLMs to reason over candidate robot behavior:
RoboMonkey samples actions from a VLA and applies a VLM-based verifier to
select among them \cite{robomonkey}, whereas Uncertainty-Aware Policy Steering
chooses whether to execute, request clarification, or obtain a corrective
intervention according to calibrated semantic and action uncertainty
\cite{ups}.

These systems do not retain candidate-specific motor evidence from the robot's
preceding contact with an object for later comparison. Our Material Explain
Module (MEM) summarizes the MIM interaction histories of two probed candidates
and provides them to a selector VLM together with the task instruction. The VLM
predicts the selected candidate and relative position/current responses, and
generates a short explanation conditioned on the same summaries, while the
action policy remains responsible for low-level position commands. We evaluate
candidate selection in real-robot trials and analyze the effect of interaction-
summary capacity using structured response prediction, explanation
consistency, and candidate-swap consistency.

\section{Method}
\label{sec:method}

\subsection{Overview}
\label{sec:method_overview}

MIM-VLA separates physical-interaction representation from task-level vision
and action generation. First, the Motor Interaction Module (MIM) encodes a
causal window of measured gripper telemetry as
$\boldsymbol{\tau}\in\mathbb{R}^{128}$. Second, the Material Explain Module
(MEM) summarizes candidate-specific token histories for a selector VLM that
compares prior interactions and produces a structured choice and explanation.
Third, the current interaction token conditions a gripper-specific SmolVLA pathway
\cite{smolvla} while
the original visual-language and arm pathways are preserved.

Training follows the same separation. MIM is pretrained with contact and phase
supervision. The interaction-memory summarizer and selector are then trained
with MIM fixed. For downstream robot control, MIM is initialized from the
pretrained encoder and jointly fine-tuned with the action pathways, while its
auxiliary classification heads remain frozen.

\subsection{Motor Interaction Module}
\label{sec:mim}

\begin{figure}[H]
    \centering
    \includegraphics[width=0.98\columnwidth]{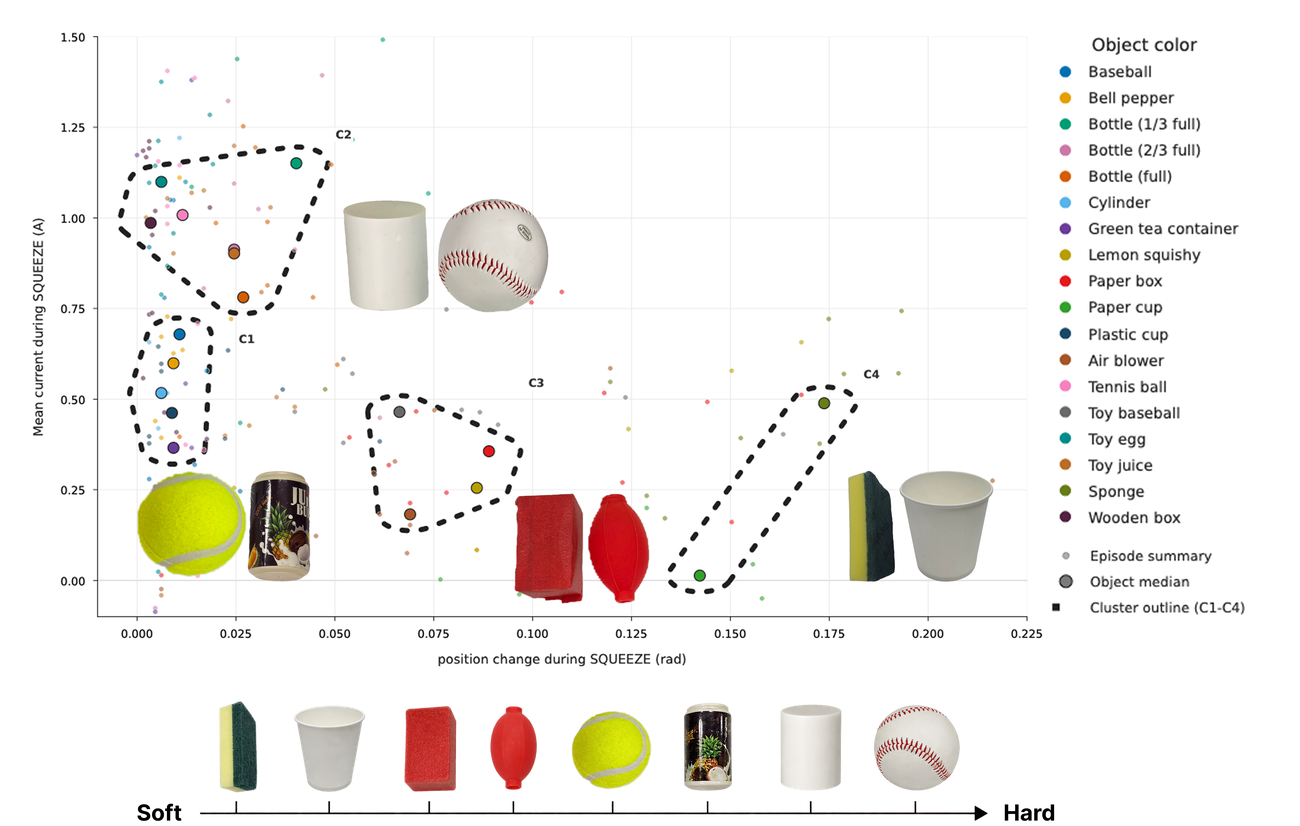}
    \caption{Qualitative motor-response validation on the MIM pretraining data.
    The 18 legend entries correspond to the objects used for MIM pretraining.
    Small points summarize individual episodes, large markers denote
    object-level medians, and dashed contours indicate recurring response
    regions (C1--C4) in gripper-position change and mean current during
    \textsc{squeeze}. Objects with similar interaction resistance exhibit
    nearby response distributions despite differences in appearance.}
    \label{fig:mim_response_clusters}
\end{figure}

MIM processes measured gripper position, velocity, and current together with
motion, torque-enable, validity, and timing signals. Commanded and target
positions, current limits, and saturation indicators are excluded from the
encoder input. Sensor and reception timestamps are used to construct causal
history windows containing only measurements available at the corresponding
observation time.

The motor-only input is deliberate. Measured position gives the gripper
aperture and cumulative closure, velocity indicates motion direction and
whether closing slows or stops, and current describes the simultaneous motor
load. Their joint temporal evolution provides the information needed for the
contact and phase distinctions supervised in this work. Excluding wrist video
also prevents the MIM token from learning an appearance shortcut. This does
not imply that telemetry replaces vision for general manipulation: camera
observations remain inputs to SmolVLM, whereas MIM is restricted to encoding
the physical response during gripper--object interaction.

Formally, for an observation issued at time $t$, MIM receives the causal
sequence
\begin{equation}
    \mathbf{X}_{t}
    = \left[\mathbf{x}_{t-L+1},\ldots,\mathbf{x}_{t}\right],
    \label{eq:mim_history}
\end{equation}
where each $\mathbf{x}$ concatenates normalized measured position, velocity,
and current, their first-order differences, motor-status bits, validity masks,
and timing information. Samples whose acquisition or reception time exceeds
$t$ are excluded even when they occur in the same logged image interval. When
a signal or temporal difference is unavailable, its validity entry is cleared
so that zero-filled normalized values cannot be mistaken for a physical zero.
This construction is important for deployment, where sensor packets arrive
asynchronously and the policy must never depend on future telemetry.

Leader--follower demonstrations cover closing, contact, additional closing,
holding, partial opening, regrasping, and release. Human reviewers annotate
binary contact and five phases: \textsc{idle}, \textsc{closing},
\textsc{squeeze}, \textsc{hold}, and \textsc{opening}. Contact requires both
fingers to visually touch the object. \textsc{Squeeze} denotes additional
closing after contact, irrespective of visible deformation, whereas
\textsc{hold} contains no additional opening or closing command. Video is used
only to generate these targets and is not a MIM input.
The MIM pretraining set contains 360 interaction episodes, with 20 episodes
collected for each of the 18 objects shown in Fig.~\ref{fig:mim_response_clusters}.

The two annotations provide complementary supervision. Contact labels identify
the transition from free motion to object interaction, whereas phase labels
describe how the response evolves before and after that transition. For
example, both a rigid object and a compliant object may be in
\textsc{squeeze}, but they can exhibit different combinations of position
advance, velocity decay, and current rise. Conversely, similar instantaneous
currents may occur during impact, sustained holding, or reopening. Supervising
the temporal encoder with both targets discourages it from reducing the
history to a single current threshold or gripper aperture.

Measurements and first-order differences are normalized with training-split
statistics and accompanied by validity masks. A learned projection with layer
normalization \cite{layernorm} and GELU \cite{gelu} maps each time step to a
feature vector. A GRU \cite{gru} encodes the sequence, and an MLP maps its final hidden state to
$\boldsymbol{\tau}$. Contact and phase heads supervise this shared
representation, as shown in Fig.~\ref{fig:mim_pretraining}:
\begin{equation}
    \mathcal{L}_{\mathrm{MIM}}
    = \lambda_{\mathrm{phase}}\mathcal{L}_{\mathrm{phase}}
    + \lambda_{\mathrm{contact}}\mathcal{L}_{\mathrm{contact}}.
    \label{eq:mim_objective}
\end{equation}
Figure~\ref{fig:mim_response_clusters} provides an empirical, qualitative
validation of the motor evidence used by MIM. Under the standardized probing
motion, higher-resistance objects show limited additional closure and stronger
current response, while more compliant objects permit greater closure with a
weaker current response. Objects with similar resistance occupy nearby regions
even when their appearance differs. The plot therefore verifies that the
measured histories contain object-dependent physical structure that a temporal
encoder can exploit. MIM is not trained from these two summary coordinates; it
receives the complete causal motor history and is supervised by the contact and
phase objectives above.

Each term is evaluated only when its target is valid. Sampling is balanced
across object families and episodes, model selection uses the validation split,
and the test split is held out. Downstream modules consume
$\boldsymbol{\tau}$ before the two classification heads rather than
their discrete predictions.

Using the pre-classification latent is essential because the auxiliary classes
are deliberately coarse. A phase prediction states whether the gripper is
closing, squeezing, holding, or opening, but does not encode the magnitude or
temporal profile of the resistance encountered within that phase. Likewise,
the binary contact target cannot distinguish early fingertip contact from a
stable load-bearing grasp. The continuous token preserves these within-class
variations and therefore supports both low-level closure adaptation and
higher-level comparison of previously explored objects. The auxiliary heads
are removed from the policy loss after pretraining; they shape the
representation without constraining downstream modules to make decisions from
the discrete labels.

\subsection{Interaction Memory and Material Reasoning}
\label{sec:interaction_memory}

MIM interaction tokens are stored separately for each candidate object. For
candidate $c\in\{A,B\}$, the interaction memory is defined as
\begin{equation}
    \mathcal{B}_{c}
    = \left\{(\boldsymbol{\tau}_{i},t_i,\mathbf{m}_{i},p_i)\right\}_{i=1}^{N_c},
    \label{eq:interaction_memory}
\end{equation}
where $\boldsymbol{\tau}_i\in\mathbb{R}^{128}$ denotes the MIM interaction token,
$t_i$ its timestamp, $\mathbf{m}_i$ the corresponding motor observation,
and $p_i$ the interaction phase. Each candidate memory is independently
constructed during exploratory grasping. The comparison interval spans from
initial contact through the subsequent closing and holding phases.

\begin{figure}[t]
    \centering
    \includegraphics[width=\columnwidth]{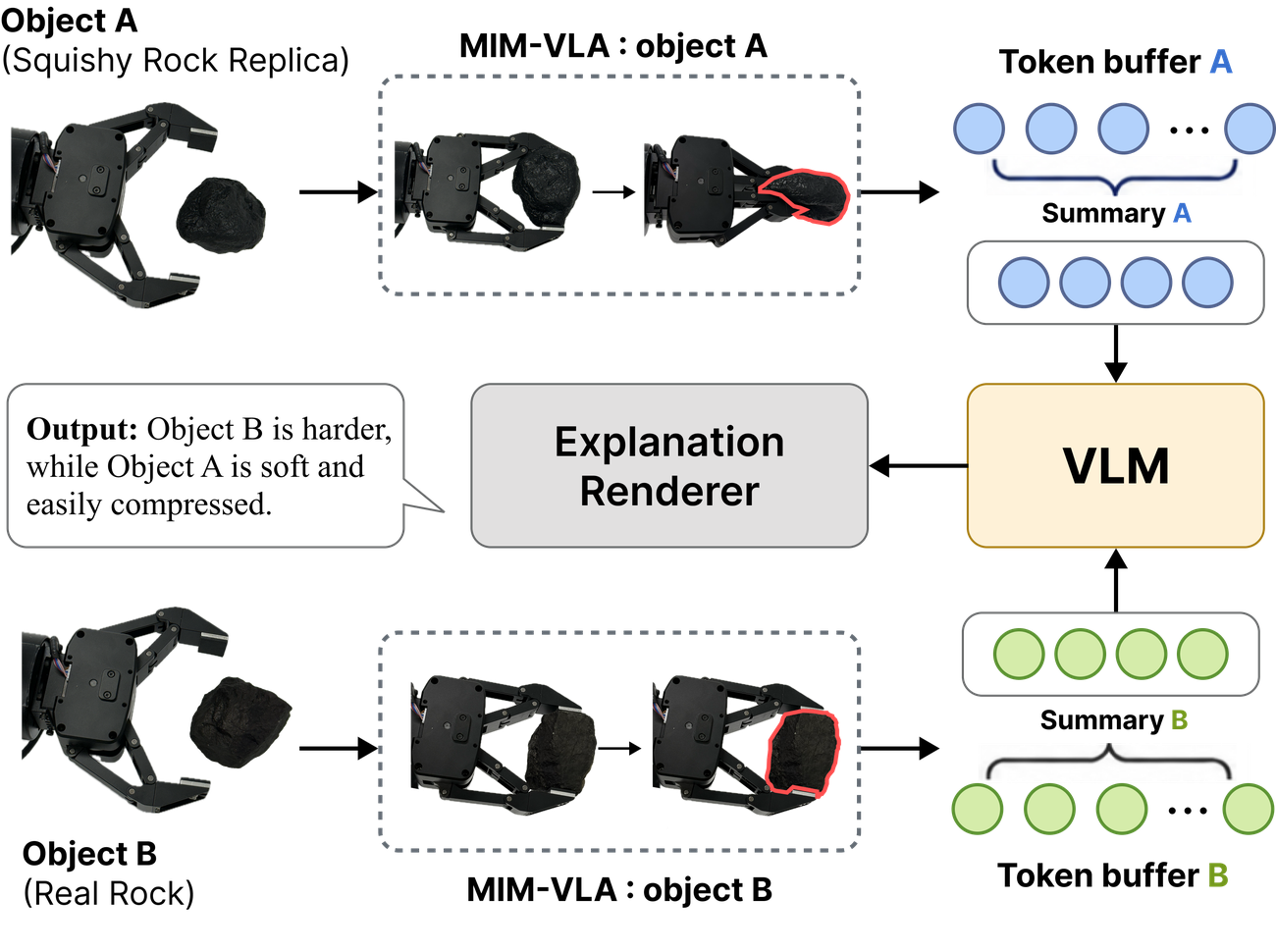}
    \caption{Candidate-specific interaction histories are summarized
    independently and compared by the Material Explain Module (MEM) to produce
    an object choice and evidence-conditioned explanation.}
    \label{fig:interaction_memory_selection}
\end{figure}

For each candidate, the MIM token sequence is combined with temporally aligned
motor features and temporal information and passed to the Temporal Summary
Module. A bidirectional temporal encoder processes the interaction sequence.
Attention, mean, max, and last-state pooling capture complementary temporal
characteristics. Each pooled context is projected to 128 dimensions, yielding
the summary representation
\begin{equation}
    \mathbf{S}_c=\left\{\mathbf{s}_c^k\right\}_{k=1}^{K},
    \qquad \mathbf{S}_c\in\mathbb{R}^{K\times128}.
    \label{eq:summary_representation}
\end{equation}
The default setting uses $K=4$, with one summary token generated from each
pooling context. Section~\ref{sec:ablations} evaluates this token capacity.

The Temporal Summary Module also predicts interaction statistics associated
with position and current responses. These auxiliary objectives encourage the
summary representation to preserve motor-response information relevant to
candidate comparison.

As summarized in Fig.~\ref{fig:interaction_memory_selection}, the summary
representations of both candidates are provided to the MEM selector VLM together with
the task instruction. A learnable projection maps the summary tokens into the
VLM embedding space. The VLM compares the two interaction representations and
produces structured predictions for candidate selection and relative
position/current responses. It also generates a natural-language explanation
conditioned on the same interaction representation:
\begin{equation}
    \begin{aligned}
    (\mathbf{S}_A,\mathbf{S}_B,x_{\mathrm{task}})
    \longmapsto{}&\bigl(\hat{y}_{\mathrm{sel}},\hat{y}_{\mathrm{pos}},\\
    &\hat{y}_{\mathrm{cur}},\hat{x}_{\mathrm{exp}}\bigr).
    \end{aligned}
    \label{eq:mem_outputs}
\end{equation}

Training proceeds in two stages. We first freeze MIM and train the Temporal
Summary Module using MIM tokens and aligned motor features. We then freeze the
Temporal Summary Module and train the VLM motor-token projection, structured
prediction heads, and LoRA parameters.

\subsection{SmolVLA Integration}
\label{sec:smolvla_integration}

We retain SmolVLA's flow-matching action formulation \cite{smolvla}
(Fig.~\ref{fig:mim_vla_architecture}). The system figure intentionally abstracts
the action block as a single expert for readability; the gripper-specific split
used during fine-tuning is detailed here. A trainable projector and token-type
embedding map $\boldsymbol{\tau}$ into the action-expert context. A
dedicated gripper expert, initialized from the pretrained action expert,
receives this token and the base arm trajectory. An asymmetric attention mask
allows the interaction token to attend to the original observation context but
blocks the reverse path.

The base expert predicts the complete joint-action sequence. The dedicated
expert predicts only the conditioned gripper channel and does not feed its
output back into the base trajectory. The executed action combines base arm
commands with the dedicated gripper command, preserving the original
position-action interface.

Policy fine-tuning uses
\begin{equation}
    \mathcal{L}_{\mathrm{policy}}
    = \mathcal{L}_{\mathrm{base}}
    + \lambda_{\mathrm{grip}}\mathcal{L}_{\mathrm{grip}},
    \label{eq:policy_objective}
\end{equation}
where the gripper term is evaluated only for valid MIM representations and
targets. The vision-language backbone remains frozen. Both action experts,
their projections, the interaction-token embedding, and the MIM encoder are
updated; gripper-loss gradients reach MIM but not the original observation
context. The pretrained contact and phase heads remain frozen and are excluded
from the policy objective.

\section{Experiments}
\label{sec:experiments}

\subsection{Experimental Setup}
\label{sec:experimental_setup}

Experiments use a 6-DoF OMY F3M follower arm with a Dynamixel two-finger
gripper, an overview RGB camera, and a wrist-mounted Intel RealSense D405.
Demonstrations are collected through an OMY leader--follower interface. Joint
states and read-only gripper telemetry are transported through ROS~2 and Zenoh.
The 100-Hz telemetry stream contains 12 command, measured-response, status, and
health channels, including position, velocity, and current. Acquisition and
reception timestamps ensure that each observation uses only motor samples
available at that time.

The gripper-only policy used in Experiment 3 is developed from 197 policy
episodes. Exporting these episodes together with 43 no-contact baselines and
three high-force diagnostic negatives yields 243 episodes and 65,512 causal
samples at 30~Hz with a 2-s, $200\times12$ motor-history window. Training uses
an RTX~5090 and batch size 64. For the
controlled Experiment 3 comparison, all deployed gripper-policy checkpoints are
taken after 10,000 updates. The policy predicts the next absolute gripper goal
position while the arm remains fixed; the Experiment 3 variants do not use
camera images as policy inputs.

Table~\ref{tab:experiment_overview} summarizes the three real-world tasks.
Experiments 1 and 2 test whether stored interaction evidence supports relative
material judgments across appearance changes and under appearance ambiguity.
Experiment 3 evaluates minimally compressive grasping of fragile objects.

\begin{table*}[t]
    \centering
    \caption{Real-world evaluation tasks. ``Unseen'' denotes held-out objects.}
    \label{tab:experiment_overview}
    \setlength{\tabcolsep}{3.5pt}
    \renewcommand{\arraystretch}{1.00}
    \scriptsize
    \begin{tabularx}{\textwidth}{@{}c
        >{\raggedright\arraybackslash}p{0.28\textwidth}
        >{\raggedright\arraybackslash}X
        >{\raggedright\arraybackslash}p{0.18\textwidth}@{}}
        \toprule
        \textbf{Exp.} & \textbf{Task} & \textbf{Evaluation protocol} & \textbf{Split} \\
        \midrule
        1 & \textbf{Cross-Appearance Hardness Comparison}
        & Sequentially probe two visually different objects, select the harder
        object, and explain the relative response.
        & 7 seen + 3 unseen pairs \\
        2 & \textbf{Appearance-Matched Physical Disambiguation}
        & Probe a real object and its appearance-matched soft replica, then
        identify the harder or genuine candidate from interaction evidence.
        & 3 matched pairs \\
        3 & \textbf{Gentle Grasping of Fragile Objects}
        & From a fixed arm pose, detect contact and establish a stable hold with
        minimal compression.
        & 8 seen + 3 unseen objects \\
        \bottomrule
    \end{tabularx}
\end{table*}

Experiments 1 and 2 compare vanilla SmolVLA \cite{smolvla}, SmolVLA with raw
current, and MIM-VLA over 20 trials per pair. Experiment 3 compares a
SmolVLA-flow gripper baseline without motor history, a temporal-current baseline
inspired by the design principles of TA-VLA \cite{tavla}, and MIM-VLA over 20
trials per object. Because TA-VLA training code was not publicly available, the
second baseline is our controlled implementation rather than an exact
reproduction. All methods use identical object placement,
camera observations, language instructions, and task-specific success
criteria.

\subsection{Material Evidence and Candidate Selection}
\label{sec:material_reasoning_evaluation}

\begin{figure}[!t]
    \centering
    \includegraphics[width=0.72\columnwidth]{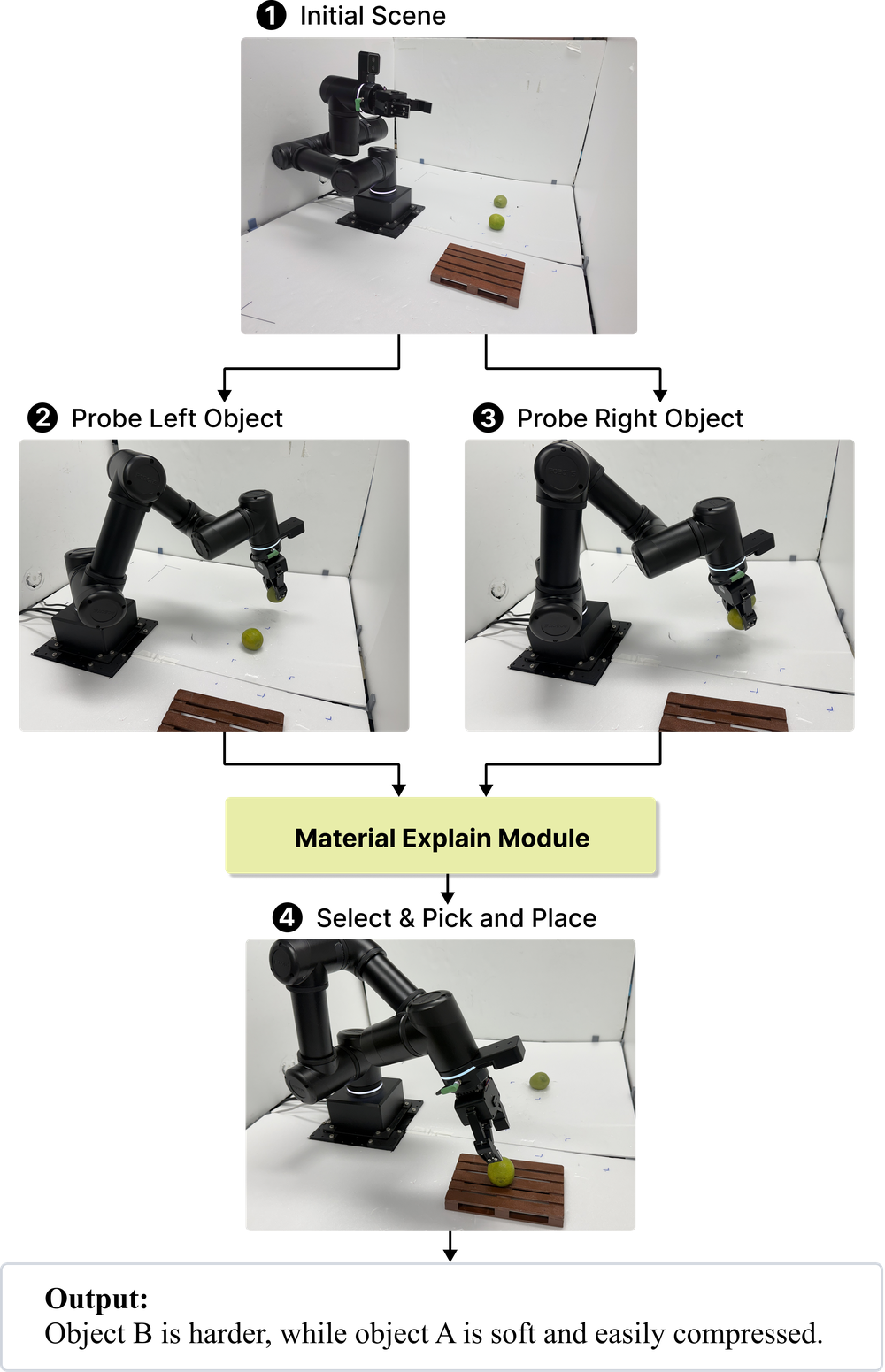}
    \caption{Shared probe--compare--select protocol for Experiments 1 and 2.
    The robot observes both candidates, probes each object, compares their
    interaction memories, and manipulates the selected candidate.}
    \label{fig:experiment12_protocol}
\end{figure}

Experiments 1 and 2 share the probe--compare--select protocol in
Fig.~\ref{fig:experiment12_protocol}. The robot probes the left object first and
then the right object with the same exploratory grasp routine; assignment of
the two candidates to the left and right positions is randomized for every
trial. MIM retains tokens from first contact through additional closing and
holding, and the Material Explain Module (MEM)
compares equal-length summaries to select the harder candidate and generate a
short explanation grounded in relative position and current responses.
We collect 1,200 training episodes in total across Experiments 1 and 2 for this
probe--compare--select pipeline.

We operationally define the target as the object exhibiting greater interaction
resistance under this standardized probing motion. Target orderings are assigned
from nominal material behavior and repeated exploratory observations, and are
qualitatively checked against closure and current responses such as those in
Fig.~\ref{fig:mim_response_clusters}. This evaluation is not intended as a
calibrated measurement of intrinsic material hardness.

Experiment 1 uses visually different objects. Its seen pairs are
Baseball/Sponge, Tennis Ball/Sponge, Plastic Cup/Paper Cup, Wooden Box/Toy Ball,
Cardboard Box/Cylindrical Plastic Model, Toy Ball/Toy Beverage Bottle, and
Wooden Box/Tennis Ball. The unseen pairs are Toy Bell Pepper/Air Blower, Golf
Ball/Empty PET Bottle, and Golf Ball/Miniature Doll. Experiment 2 instead pairs
a real object with a visually similar soft replica: Real Rock/Squishy Rock,
Real Lemon/Lemon Squishy, and Real Banana-Milk Bottle/Squishy Replica. Selection
accuracy is the primary metric for both tasks.

\begin{table*}[t]
    \centering
    \caption{Harder-target selection results for Experiments 1 and 2. Each
    object pair is evaluated over 20 trials per policy. Best results are shown
    in bold.}
    \label{tab:hardness_results}
    \setlength{\tabcolsep}{3.2pt}
    \renewcommand{\arraystretch}{1.08}
    \scriptsize
    \begin{tabularx}{\textwidth}{@{}l l >{\raggedright\arraybackslash}X
        >{\raggedright\arraybackslash}p{0.13\textwidth} c c c@{}}
        \toprule
        \textbf{Exp.} & \textbf{Split} & \textbf{Object pair}
        & \textbf{Harder target} & \textbf{\shortstack{Vanilla\\SmolVLA}}
        & \textbf{\shortstack{SmolVLA +\\Raw Current}}
        & \textbf{MIM-VLA} \\
        \midrule
        \multicolumn{7}{@{}l}{\textit{Experiment 1: Cross-Appearance Hardness Comparison}} \\
        1 & Seen & Baseball / Sponge & Baseball & 13/20 (65\%) & 9/20 (45\%) & \textbf{19/20 (95\%)} \\
        1 & Seen & Tennis Ball / Sponge & Tennis Ball & 12/20 (60\%) & 7/20 (35\%) & \textbf{19/20 (95\%)} \\
        1 & Seen & Plastic Cup / Paper Cup & Plastic Cup & 13/20 (65\%) & 6/20 (30\%) & \textbf{18/20 (90\%)} \\
        1 & Seen & Wooden Box / Toy Ball & Wooden Box & 10/20 (50\%) & 9/20 (45\%) & \textbf{16/20 (80\%)} \\
        1 & Seen & Cardboard Box / Cylindrical Plastic Model & Cylindrical Model & 13/20 (65\%) & 6/20 (30\%) & \textbf{18/20 (90\%)} \\
        1 & Seen & Toy Ball / Toy Beverage Bottle & Toy Beverage Bottle & 11/20 (55\%) & 8/20 (40\%) & \textbf{14/20 (70\%)} \\
        1 & Seen & Wooden Box / Tennis Ball & Wooden Box & 9/20 (45\%) & 4/20 (20\%) & \textbf{15/20 (75\%)} \\
        \cmidrule(lr){2-7}
        1 & \textbf{Seen avg.} & All seen pairs & --- & \textbf{81/140 (57.9\%)} & \textbf{49/140 (35.0\%)} & \textbf{119/140 (85.0\%)} \\
        \addlinespace
        1 & Unseen & Toy Bell Pepper / Air Blower & Toy Bell Pepper & 5/20 (25\%) & 4/20 (20\%) & \textbf{11/20 (55\%)} \\
        1 & Unseen & Golf Ball / Empty PET Bottle & Golf Ball & 8/20 (40\%) & 5/20 (25\%) & \textbf{13/20 (65\%)} \\
        1 & Unseen & Golf Ball / Miniature Doll & Golf Ball & 8/20 (40\%) & 5/20 (25\%) & \textbf{11/20 (55\%)} \\
        \cmidrule(lr){2-7}
        1 & \textbf{Unseen avg.} & All unseen pairs & --- & \textbf{21/60 (35.0\%)} & \textbf{14/60 (23.3\%)} & \textbf{35/60 (58.3\%)} \\
        \midrule
        \multicolumn{7}{@{}l}{\textit{Experiment 2: Appearance-Matched Physical Disambiguation}} \\
        2 & Seen & Real Rock / Squishy Rock Replica & Real Rock & 6/20 (30\%) & 3/20 (15\%) & \textbf{7/20 (35\%)} \\
        2 & Seen & Real Lemon / Lemon Squishy Replica & Real Lemon & 10/20 (50\%) & 4/20 (20\%) & \textbf{16/20 (80\%)} \\
        2 & Seen & Real Banana-Milk Bottle / Squishy Replica & Real Banana-Milk Bottle & 9/20 (45\%) & 6/20 (30\%) & \textbf{18/20 (90\%)} \\
        \cmidrule(lr){2-7}
        2 & \textbf{Average} & All Experiment 2 pairs & --- & \textbf{25/60 (41.7\%)} & \textbf{13/60 (21.7\%)} & \textbf{41/60 (68.3\%)} \\
        \midrule
        \multicolumn{2}{@{}l}{\textbf{Overall}} & All 13 object pairs & ---
        & \textbf{127/260 (48.8\%)} & \textbf{76/260 (29.2\%)} & \textbf{195/260 (75.0\%)} \\
        \bottomrule
    \end{tabularx}
\end{table*}

Table~\ref{tab:hardness_results} shows that MIM-VLA achieves 75.0\% aggregate
selection accuracy, compared with 48.8\% for vanilla SmolVLA and 29.2\% for
raw-current conditioning. The advantage remains on unseen cross-appearance
pairs (58.3\% versus 35.0\% and 23.3\%), while the appearance-matched task
reaches 68.3\%. Under this evaluation, the results are consistent with a benefit
from temporally pretrained interaction features over instantaneous raw-current
conditioning; we do not interpret the comparison as a calibrated material test.

\subsection{Gentle Grasping of Fragile Objects}
\label{sec:gentle_grasping_evaluation}

Experiment 3 isolates gripper control from arm motion. The policy is trained
on minimal-compression demonstrations spanning deformable and breakable
objects; closing stops once the object can be held without slipping. At test
time, a deterministic controller pre-opens, search-closes at low speed until
MIM contact and sustained position lag agree, executes the learned hold policy
for 10~s, and reopens. Additional learned closure is capped at 0.012~rad from
detected contact.

\begin{figure}[H]
    \centering
    \includegraphics[width=0.95\columnwidth]{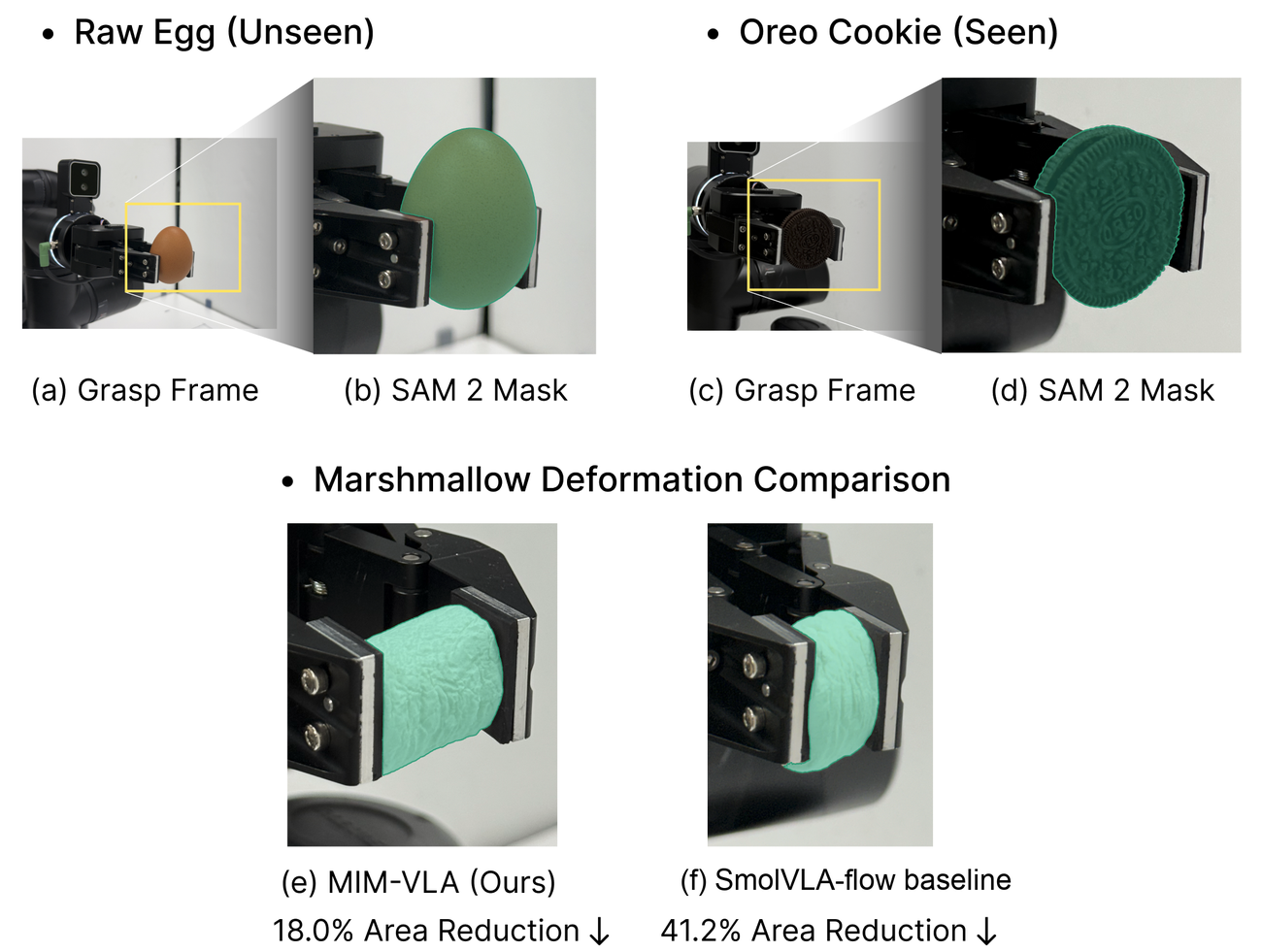}
    \caption{Qualitative evaluation of gentle grasping. Top: damage-free
    grasps of an unseen raw egg (left) and a seen Oreo cookie (right), each
    shown with its SAM~2 evaluation mask. Bottom: marshmallow deformation for
    MIM-VLA (left, 18.0\% mask-area reduction) and the SmolVLA-flow baseline
    (right, 41.2\%). SAM~2 masks are used only for evaluation.}
    \label{fig:experiment3_qualitative}
\end{figure}

Figure~\ref{fig:experiment3_qualitative} shows representative damage-free
grasps and the fixed-view SAM~2 evaluation used at maximum closure.

We record grasp success, object drop or damage, and visual deformation. A
damage-free grasp requires successful pickup, a stable 10-s hold, and release
with the object remaining intact; a drop, crack, rupture, or visible permanent
damage at or after release is counted as failure. SAM~2
\cite{sam2} is used only for evaluation: an object mask initialized before
contact is propagated through the grasp clip. The camera viewpoint and arm pose
are fixed across policies, the object remains unoccluded in the evaluation view,
and area is measured at the maximum-closure frame during the hold. Relative
mask-area reduction is
\begin{equation}
    R_A = 100\,\frac{A_0-A_{\mathrm{close}}}{A_0},
    \label{eq:mask_area_change}
\end{equation}
where $A_0$ is the pre-contact area and $A_{\mathrm{close}}$ is the mask area at
maximum closure. For each object--model pair $o$, the
$N=20$ trial-level reductions $D_{o,i}$ are summarized by
\begin{equation}
    D_o=\frac{1}{N}\sum_{i=1}^{N}D_{o,i}, \qquad
    \sigma_o=\sqrt{\frac{1}{N-1}\sum_{i=1}^{N}(D_{o,i}-D_o)^2}.
    \label{eq:deformation_mean_sd}
\end{equation}
The fixed camera and arm pose reduce viewpoint variation. SAM~2 masks are never
provided to the policy. Deformation is reported as mean $\pm$ SD; damage-free
grasping is reported as successful trials over $N$ and the corresponding rate.

\begin{table*}[t]
    \centering
    \caption{Gentle-grasp results. Deformation is mean $\pm$ SD over 20 trials
    per object and policy; lower SAM~2 mask-area reduction is better.
    Breakable-object results report successful trials and rates; higher is
    better.}
    \label{tab:fragile_results}
    \setlength{\tabcolsep}{4.2pt}
    \renewcommand{\arraystretch}{1.04}
    \scriptsize
    \begin{tabularx}{\textwidth}{@{}l >{\raggedright\arraybackslash}X c c c@{}}
        \toprule
        \textbf{Split} & \textbf{Object} & \textbf{SmolVLA flow}
        & \textbf{TA-VLA-inspired} & \textbf{MIM-VLA (Ours)} \\
        \midrule
        \multicolumn{5}{@{}l}{\textit{Deformable objects: SAM2 mask-area reduction (\%) $\downarrow$}} \\
        Seen & Marshmallow & $41.2 \pm 9.3$ & $31.0 \pm 7.6$ & \textbf{$18.0 \pm 5.4$} \\
        Seen & Empty PET Bottle & $24.8 \pm 6.7$ & $19.0 \pm 5.4$ & \textbf{$12.0 \pm 3.8$} \\
        Seen & 1/3-Filled PET Bottle & $17.6 \pm 5.1$ & $13.5 \pm 4.0$ & \textbf{$8.5 \pm 2.7$} \\
        Seen & Kitchen Sponge & $38.5 \pm 10.2$ & $30.0 \pm 8.1$ & \textbf{$17.5 \pm 5.8$} \\
        Seen & Squishy Toy & $35.7 \pm 9.4$ & $27.5 \pm 7.3$ & \textbf{$15.5 \pm 4.9$} \\
        Seen & Thin Plastic Cup & $30.6 \pm 8.2$ & $23.0 \pm 6.2$ & \textbf{$13.5 \pm 4.4$} \\
        \cmidrule(lr){1-5}
        \textbf{Seen macro avg.} & 6 objects & 31.4\% & 24.0\% & \textbf{14.2\%} \\
        Unseen & Paper Cup & $34.0 \pm 8.9$ & $25.0 \pm 7.0$ & \textbf{$15.0 \pm 5.0$} \\
        Unseen & Toy Soccer Ball & $22.0 \pm 6.0$ & $16.5 \pm 4.8$ & \textbf{$10.0 \pm 3.1$} \\
        \cmidrule(lr){1-5}
        \textbf{Unseen macro avg.} & 2 objects & 28.0\% & 20.8\% & \textbf{12.5\%} \\
        \midrule
        \multicolumn{5}{@{}l}{\textit{Breakable objects: damage-free grasp success (\%) $\uparrow$}} \\
        Seen & Cherry Tomato & 4/20 (20\%) & 7/20 (35\%) & \textbf{11/20 (55\%)} \\
        Seen & Oreo Cookie & 2/20 (10\%) & 5/20 (25\%) & \textbf{9/20 (45\%)} \\
        \cmidrule(lr){1-5}
        \textbf{Seen avg.} & 2 objects & 6/40 (15.0\%)
        & 12/40 (30.0\%) & \textbf{20/40 (50.0\%)} \\
        Unseen & Raw Egg & 1/20 (5\%) & 3/20 (15\%) & \textbf{6/20 (30\%)} \\
        \bottomrule
    \end{tabularx}
\end{table*}

Table~\ref{tab:fragile_results} shows that MIM-VLA reduces mean mask-area loss
on seen deformable objects from 31.4\% for the SmolVLA-flow baseline and 24.0\%
for the TA-VLA-inspired baseline to 14.2\%. On held-out deformable objects it obtains
12.5\%, compared with 28.0\% and 20.8\%. Damage-free success reaches 50.0\% on
seen breakable objects and 6/20 trials on the unseen raw egg. These results are
consistent with temporal motor context supporting gentler closure, while also
showing substantial room for stronger fragile-object generalization.

\subsection{Ablation Studies}
\label{sec:ablations}

This ablation is an offline evaluation of MEM using completed, held-out
interaction records. It isolates nominal candidate comparison from the
additional failures that can occur during real-robot probing and final action
execution, and therefore is not directly comparable to the end-to-end success
rates in Table~\ref{tab:hardness_results}.

We vary the number $K$ of summary tokens per candidate while keeping probing
trajectories, visual observations, and training data fixed. Table~\ref{tab:mem_summary_capacity}
shows that $K=4$ yields the best selection accuracy, explanation consistency,
and swap consistency. Although $K=2$ gives the strongest position/current F1,
four tokens provide the best overall decision and explanation performance and
are therefore used in all other experiments.

\begin{table}[t]
    \centering
    \caption{Summary-token capacity ablation. $K$ is per candidate and total
    tokens count both candidates. Results are mean $\pm$ SD over three seeds.}
    \label{tab:mem_summary_capacity}
    \setlength{\tabcolsep}{3.2pt}
    \renewcommand{\arraystretch}{1.04}
    \scriptsize
    \begin{tabular}{@{}c c c c c@{}}
        \toprule
        \textbf{$K$}
        & \textbf{\shortstack{Total tokens\\$\downarrow$}}
        & \textbf{\shortstack{Selection acc.\\$\uparrow$}}
        & \textbf{\shortstack{Position F1\\$\uparrow$}}
        & \textbf{\shortstack{Current F1\\$\uparrow$}} \\
        \midrule
        1 & \textbf{2} & $97.6 \pm 0.7$ & $78.1 \pm 3.8$ & $77.3 \pm 10.3$ \\
        2 & 4 & $98.1 \pm 0.7$ & $\mathbf{78.4 \pm 1.1}$ & $\mathbf{81.3 \pm 3.8}$ \\
        4 & 8 & $\mathbf{99.5 \pm 0.7}$ & $71.4 \pm 9.2$ & $78.0 \pm 3.2$ \\
        8 & 16 & $96.2 \pm 1.8$ & $78.0 \pm 2.4$ & $67.1 \pm 8.1$ \\
        \bottomrule
    \end{tabular}
    \vspace{2pt}

    \begin{tabular}{@{}c c c@{}}
        \toprule
        \textbf{$K$}
        & \textbf{\shortstack{Explanation consistency\\$\uparrow$}}
        & \textbf{\shortstack{Swap consistency\\$\uparrow$}} \\
        \midrule
        1 & $81.0 \pm 13.8$ & $91.9 \pm 2.9$ \\
        2 & $87.6 \pm 1.8$ & $90.0 \pm 2.0$ \\
        4 & $\mathbf{90.0 \pm 2.3}$ & $\mathbf{93.8 \pm 2.9}$ \\
        8 & $87.6 \pm 1.8$ & $89.0 \pm 1.8$ \\
        \bottomrule
    \end{tabular}
    \vspace{1pt}

    \parbox{0.98\columnwidth}{\scriptsize Each seed contains 70 test examples
    formed by applying both \textit{softer} and \textit{harder} instructions
    to 35 held-out candidate pairs.}
\end{table}

\section{Conclusion}
\label{sec:conclusion}

We presented MIM-VLA, an approach that uses onboard actuator telemetry instead
of adding a force--torque sensor or tactile array in the evaluated setup, and
converts recent gripper
position, velocity, and motor-current histories into an interaction token for
robot control and higher-level interaction-resistance reasoning. MIM preserves the robot's
position-control interface while adding temporal physical evidence through a
gripper-specific action pathway. Across the first two experiments, MIM-VLA
achieved 75.0\% harder-object selection accuracy over 13 object pairs, compared
with 48.8\% for vanilla SmolVLA and 29.2\% for raw-current conditioning. In the
gentle-grasp experiment, it reduced mean SAM~2 mask-area reduction to 14.2\% on seen
deformable objects and 12.5\% on held-out deformable objects, while improving
damage-free grasp success to 50.0\% on seen breakable objects and 30.0\% on the
unseen raw egg.

These results are consistent with the hypothesis that actuator telemetry is
more useful when modeled as a temporally supervised interaction representation
than when added as an instantaneous raw input. The same token interface also
enables candidate-specific physical evidence to be retained for subsequent
selection and explanation. Although evaluated here on pairwise resistance
comparison, MIM--MEM could extend to deformability-aware sorting, tool or grasp
strategy selection, regrasp after slip, and trial-based insertion. These remain
future directions. Current limitations include the use of one gripper platform,
the absence of calibrated force targets, sensitivity of mask-area measurements
to segmentation error, and modest success on unseen breakable objects. Future
work will extend the evaluation to broader grippers and objects, calibrate the
relationship between motor response and force, and study how MIM--MEM memories
can support multi-object, multi-step, and counterfactual physical reasoning.

\raggedbottom

{\footnotesize
\bibliographystyle{IEEEtran}
\bibliography{references_seed}
}

\end{document}